\pdfoutput=1

\documentclass[11pt]{article}

\usepackage[]{emnlp2021}

\usepackage{times}
\usepackage{latexsym}

\usepackage[T1]{fontenc}

\usepackage[utf8]{inputenc}

\usepackage{microtype}

\usepackage{graphicx}
\usepackage{multirow}

\title{Transformer Models for Text Coherence Assessment}

\author{Tushar Abhishek, Daksh Rawat, Manish Gupta\thanks{The author also works as a researcher at Microsoft}~ and Vasudeva Varma \\ Information Retrieval and Extraction Lab, IIIT Hyderabad, India \\
        tushar.abhishek@research.iiit.ac.in, daksh.rawat@students.iiit.ac.in, \\
        \{manish.gupta, vv\}@iiit.ac.in}

\begin{document}
\maketitle
\begin{abstract}
Coherence is an important aspect of text quality and is crucial for ensuring its readability. It is an essential desirable for outputs from text generation systems like summarization, question answering, machine translation, question generation, table-to-text, etc. An automated coherence scoring model is also helpful in essay scoring or providing writing feedback. 
A large body of previous work has leveraged entity-based methods, syntactic patterns, discourse relations and more recently traditional deep learning architectures for text coherence assessment. 
We hypothesize that coherence assessment is a cognitively complex task which requires deeper models and can benefit from other related tasks. Accordingly, in this paper, we propose four different Transformer-based architectures for the task: vanilla Transformer, hierarchical Transformer, multi-task learning-based model, and a model with fact-based input representation. 
Our experiments with popular benchmark datasets across multiple domains on four different coherence assessment tasks demonstrate that our models achieve state-of-the-art results outperforming existing models by a good margin.
\end{abstract}

\section{Introduction}
Coherence is a crucial metric for text quality analysis. It assimilates how well the sentences are connected and how well the document is organized. Coherent documents have a clear topic transitions that are discussed throughout the text with a smooth flow of concepts, typically in an increasing order of complexity. Ideas are first introduced in preceding sentences and are referred to later in document. Connectives are often used to assist the structure as well as smooth transitions within the document. Overall, coherence leads to better text clarity.

Coherence is vital for multiple NLP (Natural Language Processing) applications like summarization~\cite{barzilay2002inferring,parveen2016generating}, question answering~\cite{verberne2007evaluating}, machine 

 translation~\cite{xiong2019modeling,mohiuddin2020coheval}, question generation~\cite{desai2018generating}, language assessment for essay scoring~\cite{burstein2010using,somasundaran2014lexical,farag2018neural}, story generation~\cite{mcintyre2010plot}, readability assessment~\cite{pitler2010automatic,muangkammuen2020neural} and other text generation~\cite{park2015expressing,kiddon2016globally,holtzman2018learning}.

Many formal theories of coherence~\cite{grosz1995centering,mann1988rhetorical,asher2003logics} have been proposed leading to further development of various coherence models. Based on such theories, multiple text coherence models like entity-grid~\cite{barzilay2008modeling} and its extensions have been proposed. Other linguistic approaches for text coherence include coreferences, discourse relations, lexical cohesion, and syntactic features.  However, such models are incapable of handling long transitions. Also, feature engineering is decoupled from the prediction task thus limiting model performance. Recently, various models have been proposed which leverage deep learning architectures like convolutional neural networks (CNNs), recurrent neural networks (RNNs), long short-term memory networks (LSTMs). Although, Transformer~\cite{vaswani2017attention} based models have revolutionized the field of NLP, surprisingly, there is no rigorous study which investigates the effectiveness of Transformer architectures for text coherence. In this paper, we study four different kinds of Transformer architectures.

We propose models for four different coherence assessment tasks: 2-way classification, 3-way classification, sentence ordering, and coherence score prediction. For each of these tasks, we investigate effectiveness of four Transformer architectures: (1) vanilla Transformer models, (2) hierarchical Transformers with two levels, (3) multi-task learning with text coherence assessment as primary task and textual entailment as auxiliary task, and (4) fact based input representation where besides the original document, we also pass extracted facts as input. We assess the extent to which these architectures generalize to different domains and prediction tasks, establishing a new state-of-the-art (SOTA).

Overall, in this paper, we make the following main contributions. (1)
We investigate the effectiveness of four Transformer architectures. (2) We assess the extent to which the information encoded in the network generalizes to multiple domains and four prediction tasks, and demonstrate the effectiveness of our approach not only on standard sentence ordering tasks but also on more realistic task like predicting coherence of varying degrees in people's everyday writings. (3) Experiments on popular benchmark datasets (GCDC and WSJ) indicate that our proposed methods establish SOTA across multiple (task, dataset) combinations. On GCDC, our best method beats the existing SOTA by: 0.15 $F_{0.5}$ on the 2-way classification task, 3.00 absolute percent points accuracy 3-way classification task, 6.40 absolute percent points accuracy on sentence ordering task and 0.18 Spearman correlation on coherence score prediction resp. On WSJ, our best method provides an improvement of 1.28 absolute percent points on sentence ordering task. We make the code and dataset publicly available\footnote{\url{https://tinyurl.com/mry9464u}}.

\section{Related Work}
\label{sec:related}
\noindent\textbf{Entity-Grid based methods}: Discourse coherence has been studied widely using both deep learning as well as non-deep learning models.~\cite{barzilay2008modeling} proposed the entity grid model, which is based on Centering Theory~\cite{grosz1995centering}. It captures the distribution of discourse entities and transition of grammatical roles (subject, object, neither) across the sentences. Several extensions were proposed by utilising entity specific features~\cite{elsner2008coreference}, modifying ranking scheme~\cite{feng2012extending} or transforming problem into bipartite graph~\cite{mesgar2015graph}. The entity grid method as well as extensions suffer from two main drawbacks: (1) they use discrete representation for grammatical roles and features, which prevents the model from considering sufficiently long transitions due to the curse of dimensionality problem. (2) Feature engineering is decoupled from the prediction task, which limits the model's capacity to learn task-specific features. 

\noindent\textbf{Other feature engineering methods}: Besides entity grid, other linguistic approaches for text coherence include coreferences, discourse relations, lexical cohesion, and syntactic features.~\cite{elsner2008coreference} proposed a maximum-entropy based discourse-new classifier that classifies mentions of all referring expression as first mention (discourse-new) or subsequent (discourse-old) mentions.~\cite{louis2012coherence} proposed a coherence model based on syntactic patterns by assuming that sentences in a coherent discourse should share the same structural syntactic patterns. Other approaches have used syntactic patterns~\cite{louis2012coherence}, lexical cohesion~\cite{morris1991lexical,somasundaran2014lexical} or capturing topic shifts via Hidden Markov Models~\cite{barzilay2004catching}.

\noindent\textbf{Deep learning methods}: Recently, multiple deep learning approaches have been proposed.~\cite{li2014model} propose a neural framework to compute the coherence score of a document by estimating a coherence probability for each clique of $L$ sentences. \cite{li2017neural} propose generative methods to capture global topic information.~\cite{nguyen2017neural,mohiuddin2018coherence} transform entity-grid based methods into deep learning versions that obtains better result than traditional counterparts.~\cite{farag2019multi} propose a hierarchical attention model with multi-task learning objective.~\cite{xu2019cross,moon2019unified} show that modeling local coherence with discriminative models could capture both the local and the global contexts of coherence.~\cite{guz2020neural} propose an RST-Recursive model, which takes advantage of the text's RST features.~\cite{farag2020analyzing} extend some of the previous discriminative models using BERT (Bidirectional Encoder Representations from Transformers)~\cite{devlin2018bert} embeddings. Further details on some of the above mentioned methods are provided in Section~\ref{subsec:baseline}. 

Except for the last work, none of the previous methods have investigated the use of Transformer based models for text coherence modeling. Even~\cite{farag2020analyzing} simply experiment with BERT embeddings (they do not train the Transformer architecture), and that too only on one task with synthetic data. We experiment with multiple Transformer architectures. Our experiments establish a new state of the art outperforming these models by a good margin.

\section{Text Coherence Assessment Models}
\label{sec:approach}
In this section, we first discuss details of the text assessment tasks, and then present the proposed Transformer-based architectures.

\subsection{Text Coherence Assessment Tasks}
\label{subsec:text_assessment}
We propose models for four different coherence assessment tasks: (i) 3-way classification, (ii) 2-way classification, (iii) sentence ordering, (iv) coherence score prediction.

\noindent\textbf{3-way classification}: Given a document, the task is to classify it into one of the three different labels (high, medium and low) which denotes the textual coherence level of the given document.

\noindent\textbf{2-way classification}: This task is similar to the 3-way classification task, except that there are only two labels: \textit{non-coherent} or \textit{other}. 

\noindent\textbf{Sentence ordering}: For this task, every original document is assumed to be coherent. 20 random permutations (other than the original document) of sentences in the document are then obtained and labeled as negative. The dataset is created by pairing the original document and the permuted document. The task is to rank the original document higher than the permuted one in terms of coherence.

\noindent\textbf{Coherence score prediction}: A 3-point coherence score might not reflect the range of coherence that actually exists in the data. Hence, we also experiment with a coherence score prediction task. It is a regression task where, given a document, the task is to predict the degree of the textual coherence.

Note that manual judgment labels are required for 3-way classification, 2-way classification and coherence score prediction tasks. Dataset for the sentence ordering task can be generated in a synthetic manner. Also note that for all tasks, the input is a single document except for the sentence ordering task where the input is a pair of documents.
 
\subsection{Proposed Transformer based architectures}
\begin{figure}
    \centering
    \includegraphics[width=\columnwidth]{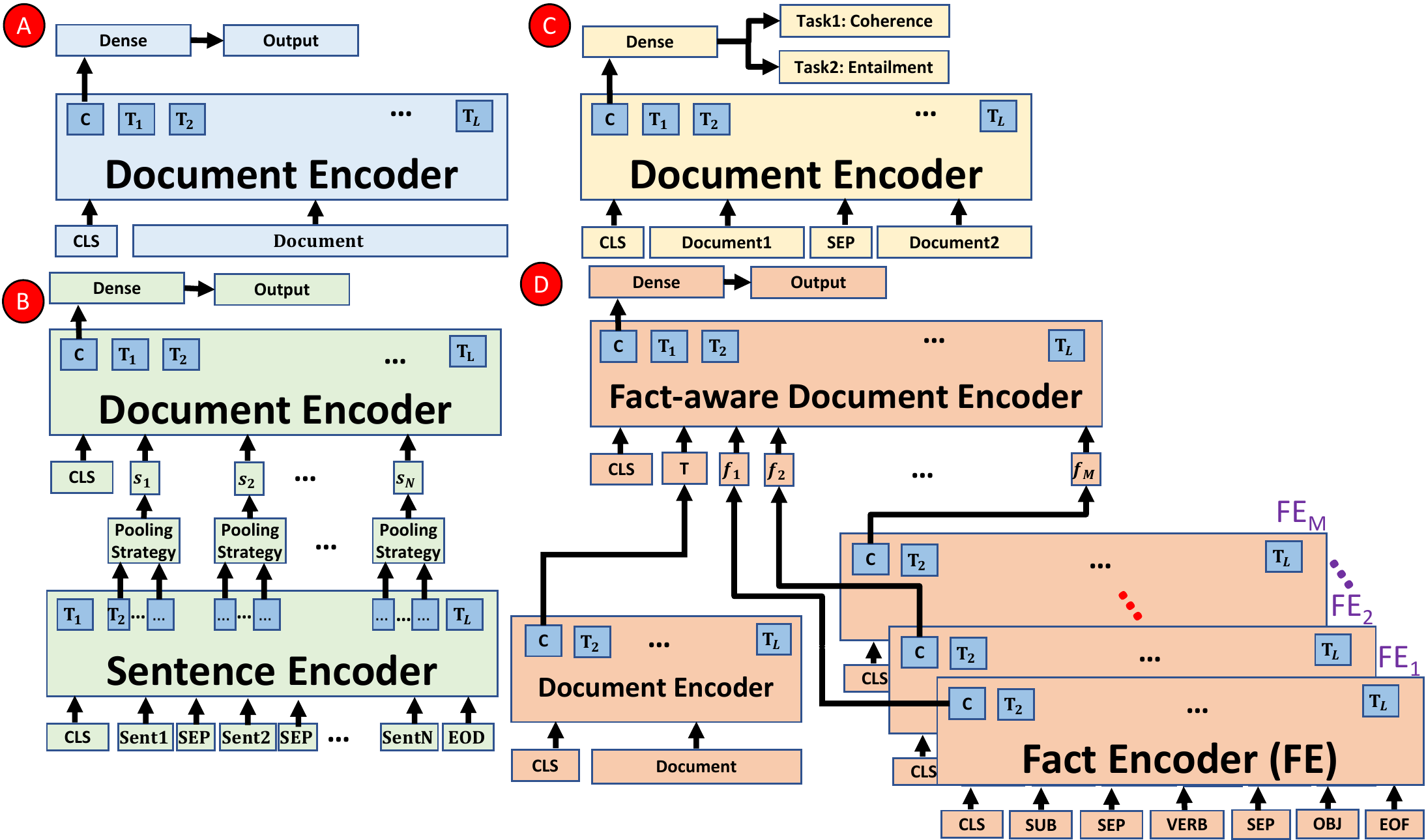}
    \caption{Architectures of various proposed methods: (A) Vanilla Transformers (B) Hierarchical Transformers (C) MTL with Transformers (D) Fact-aware Transformer}
    \label{fig:arch}
\end{figure}

Transformer~\cite{vaswani2017attention}-based deep learning models like BERT~\cite{devlin2018bert}, RoBERTa~\cite{liu2019roberta}, etc. have transformed the field of NLP in the past two years. They have been shown to be effective across a large number of NLP tasks. For each of these tasks mentioned in the previous subsection, we investigate effectiveness of four Transformer architectures: (1) vanilla Transformer models, (2) hierarchical Transformers with two levels, (3) multi-task learning with text coherence assessment and textual entailment, and (4) fact based input representation where besides the original document, we also pass extracted facts as input. We discuss details of each of these methods in the following. For each of these architectures\footnote{except for WSJ where we use Longformer. Note that Longformer has been pretrained from the RoBERTa checkpoint.} , we use RoBERTa as the basic Transformer model; we tried using BERT as well, but found RoBERTa to be better across all experiments. Fig.~\ref{fig:arch} shows the basic architecture of each of these models.

\subsubsection{Vanilla Transformer}

The simplest way of performing text coherence tasks is to feed the input text to a Transformer model as shown in Fig.~\ref{fig:arch}(A). We choose RoBERTa~\cite{liu2019roberta} which is an extension of BERT with changes to the pretraining procedure. The modifications include: (1) training the model longer, with bigger batches, over more data (2) removing the next sentence prediction objective (3) training on longer sequences, and (4) dynamically changing the masking pattern applied to the training data. RoBERTa has been shown to perform very well across many NLP tasks. RoBERTa however cannot handle very long sequences. Thus, for long sequence tasks (especially sentence ordering), we resort to Longformer~\cite{beltagy2020longformer}. Longformer has an attention mechanism that scales linearly with sequence length, making it easy to process documents of thousands of tokens or longer. Longformer's attention mechanism is a drop-in replacement for the standard self-attention and combines a local windowed attention with a task motivated global attention. 

\subsubsection{Hierarchical Transformer}
In our hierarchical Transformer architecture, we leverage a two-level hierarchical RoBERTa model for obtaining document representation as shown in Fig.~\ref{fig:arch}(B). To obtain the representation of document $D$, we use two encoders: a sentence encoder to transform each sentence in $D$ to a vector and a document encoder to learn sentence representations given their surrounding sentences as context. Both the sentence encoder and document encoder are based on the Transformer encoder. The sentence encoder takes individual sentences (separated by SEP token) as input. The outputs from each position of the last layer of the sentence encoder are pooled (separately for each sentence) using various pooling strategies like min, max, mean, sum, attention or none. In the ``none'' pooling strategy, the representation for the last sentence token is used to represent the sentence. Each sentence in the document is encoded by the sentence encoder and passed as input to the document encoder. Finally, the CLS token's transformed representation from the document encoder is fed to a dense layer with ReLU activation which is then connected to a task-specific output layer. The two encoders can either be initialized randomly or using transfer learning. 

\subsubsection{MTL with Transformers}
When multiple related prediction tasks need to be performed, multi-task learning (MTL) has been found to be very effective. Hard parameter sharing is the most commonly used approach to MTL in neural networks. It is generally applied by sharing the hidden layers between all tasks, while keeping several task-specific output layers. 

Given a pair of sentences, the textual entailment task aims to predict whether the second sentence (hypothesis) is an entailment with respect to the first one (premise) or not. Textual entailment and text coherence assessment are very similar tasks. As shown in Fig.~\ref{fig:arch}(C), we share the Transformer encoder weights across the two tasks. Task specific layers for each task are conditioned on the shared Transformer encoder. For the sentence entailment task, we form input by concatenating the hypothesis and premise with sentence separator token SEP placed between them. For both the tasks (coherence and entailment), we use a fully-connected layer with ReLU, and then a softmax output layer. The final loss is computed as a sum of the individual losses for the two tasks. 

\subsubsection{Fact-aware Transformer}
We leverage MinIE, an Open Information Extraction system~\cite{gashteovski2017minie} to generate a set of facts for each sentence. Open Information Extraction systems aim to exploit linguistic information including dependency relations in sentences to extract facts in a knowledge-agnostic manner. A fact is essentially an ordered 3-tuple (subject, verb and object)  extracted from a particular sentence. A single sentence can produce multiple facts. Consider the sentence ``They are trying to determine whether it was used to attack Steenkamp, if she used the bat in self-defense.'' Two facts that can be extracted from this sentence are  (``it'', ``was used to attack'', ``Steenkamp'') and  (``she'', ``used bat in'', ``self-defense''). Each of the three components of a fact triple can contain multiple words. 

Let the length of sequence of tokens present in document $D$ be denoted by $L$ and the number of distinct facts obtained from the document using MinIE system be denoted by $M$. As shown in Fig.~\ref{fig:arch}(D), our fact-aware Transformer method consists of three kinds of encoders: a document encoder, $M$ fact encoders ($FE_1, FE_2\ldots FE_M$) and a fact-aware document encoder. Each encoder uses a vanilla Transformer model. Document encoder and all the fact encoders share weights. Document encoder encodes the document expressed using standard sub-word tokens. Fact encoder $FE_i$ encodes the $i^{th}$ fact using the (subject, verb, object) for the fact as input delimited by SEP. Document encoder produces document representation $T$, while each of the fact encoders $FE_i$ produce fact representation $f_i$. These fact representations and the document representation form the input for the fact-aware document encoder. Finally, the representation from the last layer of the fact-aware encoder corresponding to the CLS token, is connected to a fully-connected layer with ReLU, and then a softmax output layer. 

\begin{figure}
    \centering
    \includegraphics[width=0.7\columnwidth]{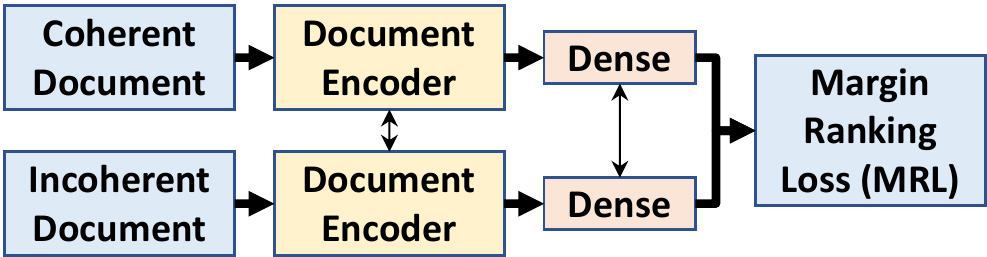}
    \caption{Overview of Siamese neural approach applied for sentence ordering task. Document encoder weights are shared. Dense layer weights are also shared.}
    \label{fig:siamese_sent_order}
\end{figure}

For all tasks except sentence ordering, we pass the document representation obtained from proposed models to a dense layer with ReLU activation which is then connected to a task-specific output layer, as shown in Fig.~\ref{fig:arch}. For sentence ordering task, we apply Siamese network~\cite{bromley1993signature} or twin neural network approach with each of our four proposed architectures, as illustrated in Fig.~\ref{fig:siamese_sent_order}. In this approach, the coherent and incoherent document representation is obtained by using any of the four document encoders (from Fig.~\ref{fig:arch}). The document encoder for the coherent as well as the incoherent document, share weights. Further, both the document representations are separately connected to a dense layer with shared weights. The outputs of the dense layers are used to calculate the margin ranking loss.

\section{Experiments}
\label{sec:experiments}
\subsection{Datasets}
We experiment with two popular benchmark datasets: Wall Street Journal (WSJ) and Grammarly Corpus of Discourse Coherence (GCDC). GCDC is a real dataset while WSJ is a synthetic dataset. We use the Recognizing Textual Entailment (RTE) dataset~\cite{wang2019superglue} for training an auxiliary task for our MTL model (2490 train and 277 validation  instances) for experiments on GCDC. For WSJ, we found MTL to perform better when we use the Multi-Genre Natural Language Inference (MNLI) dataset~\cite{N18-1101} for training the auxiliary task.

\subsubsection{WSJ} 
The WSJ portion of the Penn Treebank~\cite{elsner2008coreference,nguyen2017neural} is one of the most popular datasets for the sentence ordering task. It contains long articles without any constraint on style. Following previous work~\cite{barzilay2008modeling,nguyen2017neural}, we also use the sections 00 -- 13 for training and 14 -- 24 for testing (documents consisting of one sentence are removed). We create 20 permutations per document, making sure to exclude duplicates or versions that happen to have the same ordering of sentences as the original article. We present the basic statistics of the dataset in Table~\ref{tab:wsjStats}. 

\noindent\textbf{Sentence ordering}: We evaluate model performance on this dataset using pairwise ranking accuracy (PRA) between original text and its 20 permuted counterparts, similar to previous work. PRA calculates the fraction of correct pairwise rankings in the test data (i.e., the original coherent text should be ranked higher than its permuted non-coherent counterpart).  

\setlength{\tabcolsep}{4pt}
\begin{table}
    \centering
    \scriptsize
    \begin{tabular}{|l|l|l|l|l|}
\hline
&\#Docs&\#Synthetic Docs&Avg \#Sents&Avg \#Words\\
\hline
Train&1376&29720&21.0&529.8\\
\hline
Test&1090&21800&21.9&564.3\\
\hline
\end{tabular}
    \caption{Basic statistics of the WSJ dataset. \#Docs represents the number of original articles and \#Synthetic Docs represents the number of original articles and their permuted versions.}
    \label{tab:wsjStats}
\end{table}

\subsubsection{GCDC}
The GCDC dataset contains emails and reviews written with varying degrees of proficiency and care~\cite{lai2018discourse}. The WSJ dataset contains documents that have been professionally written and extensively edited. In contrast to WSJ, the GCDC dataset contains writing from non-professional writers in everyday contexts. Rather than using permuted or machine generated texts as examples of low coherence, GCDC has real sentences in which people try but fail to write coherently. GCDC is a corpus that contains texts from four domains, covering a range of coherence, each annotated with a document-level coherence score. Specifically, the dataset contains texts from four domains: Yahoo online forum posts, emails from Hillary Clinton's office, emails from Enron and Yelp business reviews.  We present the basic statistics of the dataset in Table~\ref{tab:gcdcStats}.

\begin{table}
    \centering
    \scriptsize
    \begin{tabular}{|l|l|p{0.3in}|p{0.3in}|p{0.3in}|p{0.6in}|l|}
\hline
&\#Docs&Avg \#Words&Avg \#Sents&Vocab Size&L, M, H Instances (\%)&NC(\%)\\
\hline
Yahoo&1200&162.1&7.5&13235&46.6,17.4,37.0&30.0\\
\hline
Clinton&1200&189.0&6.6&15564&28.2,20.6,51.2&16.6\\
\hline
Enron&1200&196.2&7.7&13694&29.9,19.4,50.7&18.4\\
\hline
Yelp&1200&183.1&7.5&12201&27.1,21.8,51.1&14.8\\
\hline
    \end{tabular}
    \caption{Basic statistics of the GCDC dataset. For each of these domains, a fixed split of 1000 and 200 was used for train and test respectively as specified in [Lai and Tetreault, 2018]. NC=Non-coherent, L=low, M=medium, H=High.}
    \label{tab:gcdcStats}
\end{table}

\noindent\textbf{3-way classification}: For each of these domains, a fixed split of 1000 and 200 was used for train and test respectively as specified in~\cite{lai2018discourse}. Of the 1000 documents, we use 200 documents for validation and remaining 800 for training. 
For our experiments, we use the consensus rating of the expert scores as calculated by~\cite{lai2018discourse}, and train our models for all the four tasks. To evaluate model performance, we use 3-way classification accuracy.

\noindent\textbf{2-way classification}: A text is labeled as non-coherent if at least two expert annotators judged the text to be low coherence, or labeled as ``other'' otherwise. For 2-way classification task, we report the $F_{0.5}$ score of low coherence class where precision is emphasized twice as much as recall. This is in line with evaluation standards in writing feedback applications~\cite{ng2014conll} and also in previous text coherence studies~\cite{lai2018discourse}.

\noindent\textbf{Sentence ordering}: Similar to \cite{lai2018discourse} we created the synthetic dataset for this task in the same way as for WSJ. Since the sentence ordering task assumes well-formed texts, we use
only the high coherence texts. Statistics of the GCDC dataset wrt this task are shown in Table~\ref{tab:gcdcStats2}. We use PRA as the metric for this task as well.

\begin{table}
    \centering
    \scriptsize
    \begin{tabular}{|l|l|l|p{0.7in}|p{0.6in}|}
\hline
&\#Train Docs&\#Test Docs&\#Train Synthetic Docs&\#Test Synthetic Docs\\
\hline
Yahoo&369&76&5905&1520\\
\hline
Clinton&511&111&8150&2220\\
\hline
Enron&507&88&8105&1760\\
\hline
Yelp&511&108&8180&2160\\
\hline
    \end{tabular}
    \caption{Basic statistics of the GCDC dataset for the sentence ordering task. We used the same train and test split as specified in [Lai and Tetreault, 2018].}
    \label{tab:gcdcStats2}
\end{table}

\noindent\textbf{Coherence Score Prediction}: For this task, gold score is the mean of 3 expert rater judgments (low coherence = 1, medium = 2, high = 3). We report Spearman's rank correlation coefficient between the gold scores and the predicted coherence scores.



\subsection{Baselines}
\label{subsec:baseline}
We experiment with the following baselines. While \textbf{Flesch-Kincaid grade level}~\cite{kincaid1975derivation} is a readability measure, previous work has treated readability and text coherence as overlapping tasks~\cite{barzilay2008modeling,mesgar2015graph}. For coherence classification, we search over the grade level scores on the training data and select thresholds that result in the highest accuracy.  \textbf{Entity grid (EGRID)}~\cite{barzilay2008modeling} builds an  entity grid which is a matrix that tracks entity mentions over sentences. Random forest classifier is trained over features extracted from  entity grid. 
\textbf{CNN-Egrid}~\cite{nguyen2017neural} is a local coherence model that employs a CNN that operates over the entity grid representation. 
\textbf{Lexicalized CNN-Egrid}~\cite{mohiuddin2018coherence} extends CNN-Egrid with lexical information about the entities.  In \textbf{Local Coherence Model (LC)}~\cite{li2014model}, sentences are encoded with a recurrent or recursive layer and a filter of weights is applied over each window of sentence vectors to extract  scores that are aggregated to calculate overall document coherence score.  \textbf{Paragraph sequence (PARSEQ)}~\cite{lai2018discourse} contains three stacked LSTMs to represent sentence, paragraph and document.  \textbf{Hierachical LSTM}~\cite{farag2019multi} is very similar to PARSEQ, but with attention and uses BiLSTMs.  \textbf{Coh+GR}~\cite{farag2019multi} extends Hierachical LSTM by training it to predict word-level labels indicating the predicted grammatical role (GR) type at the bottom layers of the network, along with the document-level coherence score. 
\textbf{Coh+SOX}~\cite{farag2019multi} is same as Coh+GR where, for each word, we only predict subject (S), object (O) and `other' (X) roles.  \textbf{Seq2Seq}~\cite{li2017neural} consists of two LSTM generative language models and uses the difference between conditional log likelihood of a sentence given its preceding/succeeding context, and the marginal log likelihood of the current/next sentence to assess coherence. 
\textbf{Local Coherence Discriminator (LCD-G)}~\cite{xu2019cross} uses averaged GloVe embeddings as the sentence representations. A representation for two consecutive sentences is then computed by concatenating the output of a set of linear transformations applied to the two sentences. This is fed to a dense layer and used to predict a local coherence score. \textbf{LCD-L}~\cite{xu2019cross} is similar to LCD-G, but applies max-pooling on the hidden state of the language model to get the sentence representation. 
\textbf{Coh+GR\_BERT}~\cite{farag2020analyzing} is similar to Coh+GR, except that BERT embeddings are used instead of GloVe embeddings as input to BiLSTMs. 
\textbf{LCD\_BERT}~\cite{farag2020analyzing} is similar to LCD-G but uses averaged BERT (instead of GloVe) embeddings as the sentence representations.  \textbf{Unified}~\cite{moon2019unified} uses a combination of LSTMs and CNNs.   \textbf{Sentence averaging (SENTAVG)}~\cite{lai2018discourse} ignores sentence order and is similar to 2-level (sentence and document) hierarchical LSTM.

\subsection{Experimental Settings}
All experiments were run on a machine equipped with four 32GB V100 GPUs. For all Transformer models, we use 12-layer models, and embedding layer was frozen except for the sentence ordering task on WSJ. For Hierarchical Transformer models, we used pretrained model for the sentence encoder and a randomly initialized RoBERTa for document encoder. For fact-aware Transformer models, we used pretrained model for the fact encoders and document encoder, and a randomly initialized RoBERTa for fact-aware document encoder. 
For all experiments we run 10 epochs, weight decay of 0.01 and use a dropout of 0.1. We use Adam optimizer for all the experiments, except for WSJ experiments where we use AdamW. For all the baseline models, we report results from their original papers. For all of our models, the reported results are the mean of 10 runs with different random seeds. When we use margin ranking loss, where margin is set to 1. For MTL based Transformers,  categorical cross entropy loss was used for the auxiliary task. For Longformer, we fixed max sequence length to 2048. For RoBERTa, we fixed it to 512.

\subsection{Results}
Tables~\ref{tab:sentenceOrderingWSJ} to~\ref{tab:twoWayGCDC} show the results across all the four text coherence tasks for WSJ and GCDC datasets. Broadly we observe that our proposed Transformer-based methods significantly outperform baselines, establishing a new SOTA across all tasks. We also experimented with a fact-aware hierarchical MTL model (or a combined model) but observed that the combination does not lead to any better results. 

\noindent\textbf{Sentence ordering results}: 
Tables~\ref{tab:sentenceOrderingWSJ} and~\ref{tab:sentenceOrderingGCDC} show results for the sentence ordering task for WSJ and GCDC datasets respectively. We make the following observations for WSJ results (Table~\ref{tab:sentenceOrderingWSJ}): (1) Fact aware transformer outperforms hierarchical model as it can incorporate the factual information flow (subject in discourse) in addition to textual information which helps it to correctly determine the coherent  sentences. (2) MTL with Transformers outperforms other variants as the auxiliary task helps in better generalization over test set.

\begin{table}[!ht]
    \centering
    \scriptsize
    \begin{tabular}{|l|l|l|}
    \hline
&Model&PRA\\
\hline
\hline
\multirow{9}{*}{\rotatebox{90}{Baselines}}&LC&74.10\\
\cline{2-3}
&PARSEQ&74.10\\
\cline{2-3}
&Seq2Seq&86.95\\
\cline{2-3}
&CNN-Egrid&88.69\\
\cline{2-3}
&Unified (ELMo)&93.19\\
\cline{2-3}
&Coh+GR&93.20\\
\cline{2-3}
&LCD-L&95.49\\
\cline{2-3}
&Coh+GR\_BERT&96.10\\
\cline{2-3}
&LCD\_BERT&97.10\\
\hline
\multirow{3}{*}{\rotatebox{90}{Ours}}&Vanilla Transformer&97.34\\
\cline{2-3}
&Hierarchical Transformer&97.55\\
\cline{2-3}
&Fact-aware Transformer&97.81\\
\cline{2-3}
&MTL with Transformers&\textbf{98.38}\\
\hline
    \end{tabular}
    \caption{Sentence ordering task PRA results on WSJ}
    \label{tab:sentenceOrderingWSJ}
\end{table}
We make the following observations for GCDC results (Table~\ref{tab:sentenceOrderingGCDC}): (1) RTE and coherence tasks help improve accuracy of each other and this leads to best results using MTL with Transformers. (2) Among the baselines, LCD models perform better with respect to other baselines. This shows that local coherence is a good indicator of global document coherence. (3) LCD\_BERT is better than all other LCD models indicating that using just Transformer embeddings can also be helpful. (4) Models built using training data across all the four domains lead to better results compared to models trained on individual domain data. (5) While baseline methods work reasonably well on other three domains, their accuracy on Yahoo is very low. Our method has massive gains on Yahoo compared to baselines.

\setlength{\tabcolsep}{2.5pt}
\begin{table}[!ht]
    \centering
    \scriptsize
    \begin{tabular}{|l|l|l|l|l|l||l|}
\hline
&Model&Yahoo&Clinton&Enron&Yelp&Average\\
\hline
\hline
\multirow{6}{*}{\rotatebox{90}{Baselines}}&CNN-Egrid&54.8&75.5&73.1&58.7&65.5\\
\cline{2-7}
&EGRID&55.9&78.2&77.4&62.9&68.6\\
\cline{2-7}
&PARSEQ&58.3&88.0&87.1&74.2&76.9\\
\cline{2-7}
&LCD-L&82.1&87.7&87.1&94.8&87.9\\
\cline{2-7}
&LCD-G&76.3&93.5&95.3&92.9&89.5\\
\cline{2-7}
&LCD\_BERT&83.6&94.3&96.7&94.4&92.3\\
\hline
\multirow{7}{*}{\rotatebox{90}{Ours}}&Fact-aware Transformer&84.2&98.7&83.2&84.1&87.6\\
\cline{2-7}
&MTL with Transformers&88.0&94.2&95.0&98.6&94.0\\
\cline{2-7}
&Vanilla Transformer&88.2&94.4&95.2&98.1&94.0\\
\cline{2-7}
&Hierarchical Transformer (all)&90.7&96.9&97.8&98.7&96.0\\
\cline{2-7}
&Vanilla Transformer (all)&91.0&97.1&98.1&98.9&96.3\\
\cline{2-7}
&Hierarchical Transformer&94.1&98.2&99.1&\textbf{99.8}&97.8\\
\cline{2-7}
&MTL with Transformers (all)&\textbf{96.3}&\textbf{99.1}&\textbf{99.4}&\textbf{99.8}&\textbf{98.7}\\
\hline
    \end{tabular}
    \caption{Sentence ordering PRA results on GCDC. ``(all)'' indicates that training data across all four domains was used.}
    \label{tab:sentenceOrderingGCDC}
\end{table}


\noindent\textbf{3-way classification results}: Table~\ref{tab:threeWayGCDC} shows 3-way classification results on GCDC. We make the following observations: (1) While MTL with Transformers leads to best results on average across all domains, there is no unique winner for individual domains. Still, our proposed methods outperformed the baselines. (2) Re-ordering of associated facts helps the fact-aware Transformer model in identifying the change in sentence order (in WSJ synthetic data) but seems ineffective on realistic  (GCDC) documents. (3) Out of the three gold coherence labels (low, medium, high) all the models have difficulty in correctly classifying documents of medium level coherence, which can be attributed to the smaller number of training examples for that particular class.

\setlength{\tabcolsep}{2.5pt}
\begin{table}[!ht]
    \centering
    \scriptsize
    \begin{tabular}{|l|l|l|l|l|l||l|}
\hline
&Model&Yahoo&Clinton&Enron&Yelp&Average\\
\hline
\hline
\multirow{10}{*}{\rotatebox{90}{Baselines}}&EGRID+coref&41.5&48.0&47.0&49.0&46.4\\
\cline{2-7}
&EGRAPH+coref&42.5&55.0&44.0&54.0&48.9\\
\cline{2-7}
&Lexicalized CNN-Egrid+coref&51.0&56.6&44.7&54.0&51.6\\
\cline{2-7}
&Flesch-Kincaid grade level&43.5&56.0&52.5&55.0&51.8\\
\cline{2-7}
&Coh+SOX&50.5&58.5&51.0&-&53.3\\
\cline{2-7}
&Hierachical LSTM&55.0&59.0&50.5&-&54.8\\
\cline{2-7}
&PARSEQ&54.9&60.2&53.2&54.4&55.7\\
\cline{2-7}
&LC&53.5&61.0&54.4&-&56.3\\
\cline{2-7}
&PARSEQ (all)&58.5&61.0&53.9&56.5&57.5\\
\cline{2-7}
&Coh+GR&56.0&62.0&56.0&-&58.0\\
\hline
\multirow{7}{*}{\rotatebox{90}{Ours}}&Fact-aware Transformer&58.4&64.3&53.8&58.1&58.7\\
\cline{2-7}
&Vanilla Transformer (all)&58.1&63.9&55.3&57.6&58.7\\
\cline{2-7}
&MTL with Transformers&58.9&\textbf{66.9}&55.1&56.6&59.4\\
\cline{2-7}
&Vanilla Transformer&59.3&65.6&56.2&57.4&59.6\\
\cline{2-7}
&Hierarchical Transformer&60.7&66.0&55.4&\textbf{59.2}&60.3\\
\cline{2-7}
&Hierarchical Transformer (all)&60.9&66.5&56.3&58.5&60.6\\
\cline{2-7}
&MTL with Transformer (all)&\textbf{62.2}&66.5&\textbf{56.4}&58.9&\textbf{61.0}\\
\hline
    \end{tabular}
    \caption{3-way classification accuracy results on GCDC. ``(all)'' indicates that training data across all four domains was used.}
    \label{tab:threeWayGCDC}
\end{table}

\noindent\textbf{Coherence Score Prediction results}: Table~\ref{tab:cohScoreGCDC} shows Coherence Score Prediction results on GCDC. We observe that (1) textual entailment as auxiliary task helps MTL with Transformer to achieve best results. (2) Although PARSEQ and Hierarchical Transformers are both hierarchical in nature, Transformer-based training in Hierarchical Transformers helps them outperform the PARSEQ model.

\begin{table}[!ht]
    \centering
    \scriptsize
    \begin{tabular}{|l|l|l|l|l|l||l|}
\hline
&Model&Yahoo&Clinton&Enron&Yelp&Average\\
\hline
\hline
\multirow{5}{*}{\rotatebox{90}{Baselines}}&EGRID&0.110&0.146&0.168&0.121&0.136\\
\cline{2-7}
&CNN-Egrid&0.204&0.251&0.258&0.104&0.204\\
\cline{2-7}
&Flesch-Kincaid grade level&0.089&0.323&0.244&0.200&0.214\\
\cline{2-7}
&SENTAVG&0.466&0.505&0.438&0.311&0.430\\
\cline{2-7}
&PARSEQ&0.519&0.448&0.454&0.329&0.438\\
\hline
\multirow{7}{*}{\rotatebox{90}{Ours}}&Fact-aware Transformer&0.638&0.610&0.474&0.426&0.537\\
\cline{2-7}
&Hierarchical Transformer&0.662&0.599&0.512&0.452&0.556\\
\cline{2-7}
&MTL with Transformers&0.655&0.666&0.561&0.468&0.588\\
\cline{2-7}
&Vanilla Transformer&0.668&0.653&0.560&0.484&0.591\\
\cline{2-7}
&Hierarchical Transformer (all)&0.671&0.673&0.587&\textbf{0.499}&0.608\\
\cline{2-7}
&Vanilla Transformer (all)&\textbf{0.690}&0.689&\textbf{0.599}&0.493&0.618\\
\cline{2-7}
&MTL with Transformers (all)&0.684&\textbf{0.709}&0.598&0.495&\textbf{0.622}\\
\hline
    \end{tabular}
    \caption{Coherence Score Spearman correlation Prediction results on GCDC. ``(all)'' indicates that training data across all four domains was used.}
    \label{tab:cohScoreGCDC}
\end{table}

\noindent\textbf{2-way classification results}: Table~\ref{tab:twoWayGCDC} shows 2-way classification results on GCDC. (1) Vanilla Transformer trained on individual domain datasets work best on average. For Yahoo and Yelp, training on all domains is beneficial. (2) On average, our best method leads to $F_{0.5}$ of 0.506 as compared to the existing SOTA (SENTAVG) of 0.351. Thus, the improvements are significant. 

\begin{table}[!ht]
    \centering
    \scriptsize
    \begin{tabular}{|l|l|l|l|l|l||l|}
\hline
&Model&Yahoo&Clinton&Enron&Yelp&Average\\
\hline
\hline
\multirow{5}{*}{\rotatebox{90}{Baselines}}&EGRID&0.258&0.260&0.294&0.161&0.243\\
\cline{2-7}
&CNN-Egrid&0.360&0.238&0.279&0.169&0.262\\
\cline{2-7}
&Flesch-Kincaid grade level&0.283&0.255&0.341&0.197&0.269\\
\cline{2-7}
&PARSEQ&0.447&0.296&0.373&0.112&0.307\\
\cline{2-7}
&SENTAVG&0.481&0.332&0.393&0.199&0.351\\
\hline
\multirow{7}{*}{\rotatebox{90}{Ours}}&Hierarchical Transformer (all)&0.566&0.407&0.369&0.272&0.404\\
\cline{2-7}
&Fact-aware Transformer&0.574&0.416&0.471&0.237&0.425\\
\cline{2-7}
&MTL with Transformers (all) &0.587&0.415&0.401&\textbf{0.406}&0.452\\
\cline{2-7}
&Vanilla Transformer (all) &\textbf{0.591}&0.491&0.397&0.381&0.465\\
\cline{2-7}
&MTL with Transformers&0.584&0.471&0.511&0.300&0.467\\
\cline{2-7}
&Hierarchical Transformer&0.584&0.520&0.518&0.342&0.491\\
\cline{2-7}
&Vanilla Transformer&0.574&\textbf{0.525}&\textbf{0.546}&0.377&\textbf{0.506}\\
\hline
    \end{tabular}
    \caption{2-way classification $F_{0.5}$ results on GCDC. ``(all)'' indicates that training data across all four domains was used.}
    \label{tab:twoWayGCDC}
\end{table}

\section{Conclusion}
\label{sec:conclusion}
In this paper, we investigate the efficacy of four different transformer architectures: vanilla, hierarchical, MTL with transformer, and fact-aware transformer on four different coherence assessment tasks: 2-way classification, 3-way classification, sentence ordering, and coherence score prediction. We observe that all the transformer based approaches outperforms existing models. It indicate that deeper models like transformer captures textual coherence signal well. 
Our work also shows that inductive transfer between MTL-based transformer models' tasks: textual coherence assessment and textual entailment, helps it achieve the best performance for most coherence assessment tasks.  

In the future, we plan to extend this work to evaluate the text coherence in an open domain setting. We also plan to evaluate the impact of coherence estimation on downstream applications like essay evaluation and summarization.
\bibliography{references}
\bibliographystyle{acl_natbib}
\end{document}